\documentclass[11pt]{article}
\usepackage{graphicx}
\usepackage{longtable,lscape}
\usepackage{amsthm}
\usepackage{amsfonts}
\usepackage{amsmath}
\usepackage{bbm}
\usepackage{float}
\usepackage{url}

\newcommand{\captionfonts}{\footnotesize}
\makeatletter 
\long\def\@makecaption#1#2{%
\vskip\abovecaptionskip
\sbox\@tempboxa{{\captionfonts #1: #2}}%
\ifdim \wd\@tempboxa >\hsize
{\captionfonts #1: #2\par}
\else
\hbox to\hsize{\hfil\box\@tempboxa\hfil}%
\fi
\vskip\belowcaptionskip}
\makeatother

\title{{\bf On the Foundations of the Brussels Operational-Realistic Approach to Cognition}}

\author{Diederik Aerts\footnote{Center Leo Apostel for Interdisciplinary Studies and Department of Mathematics, Brussels Free University, Krijgskundestraat 33, 1160 Brussels (Belgium). Email address: \emph{diraerts@vub.ac.be}} \quad
Massimiliano Sassoli de Bianchi\footnote{Laboratorio di Autoricerca di Base, 6914 Lugano (Switzerland). E-Mail address: \emph{autoricerca@gmail.com}}
\quad Sandro Sozzo\footnote{School of Management and Institute IQSCS, University Road, LE1 7RH Leicester (United Kingdom). Email address: \emph{ss831@le.ac.uk}}}

\begin{document}

\maketitle

\begin{abstract}
\noindent
The scientific community is becoming more and more interested in the research that applies the mathematical formalism of quantum theory to model human decision-making. In this paper, we provide the theoretical foundations of the quantum approach to cognition that we developed in Brussels. These foundations rest on the results of two decade studies on the axiomatic and operational-realistic approaches to the foundations of quantum physics. The deep analogies between 
the foundations of physics and cognition lead us to investigate the validity of quantum theory as a general and unitary framework for cognitive processes, and the empirical success of the Hilbert space models derived by such investigation provides a strong theoretical confirmation of this validity. However, two situations in the cognitive realm,  `question order effects' and `response replicability', indicate that even the Hilbert space framework could be insufficient to reproduce the collected data. This does not mean that the mentioned operational-realistic approach would be incorrect, but simply that a larger class of measurements would be in force in human cognition, so that an extended quantum formalism  
may be needed to deal with all of them. As we will explain, the recently derived `extended Bloch representation' of quantum theory (and the associated `general tension-reduction' model) precisely provides such extended formalism, 
while remaining within the same unitary interpretative framework. 

\end{abstract}

\section{Introduction\label{intro}}
A fundamental problem in cognition concerns the identification of the principles guiding human decision-making. Identifying the mechanisms of decision-making would indeed have manifold implications, from psychology to economics, finance, politics, philosophy, and computer science. In this regard, the predominant theoretical paradigm rests on a classical conception of logic and probability theory. According to this paradigm, people take decisions by following the rules of Boole's logic, while the probabilistic aspects of these decisions can be formalized by Kolmogorov's probability theory \cite{k1933}. However, increasing experimental evidence on conceptual categorization, probability judgments and behavioral economics confirms that this classical conception is fundamentally problematical, in the sense that the cognitive models based on these mathematical structures are not capable of capturing how people concretely take decisions in situations of uncertainty.

In the last decade, an alternative scientific paradigm has caught on which applies a different modeling scheme. The research that uses the mathematical formalism of quantum theory to model situations and processes in cognitive science is becoming more and more accepted in the scientific community, having attracted the interest of renowned scientists, funding institutions, media and popular science. And, quantum models of cognition showed to be more effective than traditional modeling schemes to describe situations like the `Guppy effect', the `combination problem', the `prisoner's dilemma', the `conjunction and disjunction fallacies', `similarity judgments', the `disjunction effect', `violations of the Sure-Thing principle', `Allais', `Ellsberg' and `Machina paradoxes' (see, e.g., \cite{ga2002,ag2005a,ag2005b,a2009a,pb2009,k2010,bpft2011,bb2012,abgs2013,ags2013,hk2013,pb2013,wbap2013,ast2014}).

There is a general acceptance that the use of the term `quantum' is not directly related to physics, neither this research in `quantum cognition' aims to unveil the microscopic processes occurring in the human brain. The term `quantum' rather refers to the mathematical structures that are applied to cognitive domains. The scientific community engaged in this research does not instead have a shared opinion on how and why these quantum mathematical structures should be employed in human cognition. Different hypotheses have been put forward in this respect. Our research team in Brussels has been working in this domain since early nineties, providing pioneering and substantial contributions to its growth, and we think it is important  to expose the epistemological foundations of the quantum theoretical approach to cognition we developed in these years. This is the main aim of the present paper. 

Our approach was inspired by a two decade research on the mathematical and conceptual foundations of quantum physics, quantum probability and the fundamental differences between classical and quantum structures \cite{a82,a86,p1989,a99a}. We followed an axiomatic and operational-realistic approach to quantum physics, in which we investigated how the mathematical formalism of quantum theory in Hilbert space can be derived from more intuitive and physically justified axioms, directly connected with empirical situations and facts. This led us to elaborate a `State Context Property' (SCoP) formalism, according to which any physical entity is expressed in terms of the operationally well defined notions of `state', `context' and `property', and functional relations between these notions. If suitable axioms are imposed to such a SCoP structure, then one obtains a mathematical representation
that is isomorphic to a Hilbert space over complex numbers (see, e.g., \cite{bc81}).

Let us shortly explain the 
`operational-realistic' connotation characterizing
our approach, because doing so we can easily point out its specific strength, and the reason why it introduces an essentially new element to the domain of psychology. `Operational' stands for the fact that all fundamental elements in the formalism  are directly linked to the measurement settings and operations that are performed in the laboratory of experimentation. `Realistic' means that we introduce 
in an operational way
the notion of `state of an entity', considering such a `state' as representing an aspect of the
reality of the considered entity at a specific moment or during a specific time-span. Historically, the notion of `state of a physical entity' was the `easy' part of the physical theories that were the predecessors of quantum theory, and it was the birth of quantum theory that forced physicists to take also 
seriously
the role of measurement and hence the value of an operational approach. The reason is that `the reality of a physical entity' was considered to be a simple and straightforward notion in classical physics and hence the
`different modes of reality of  a
same physical entity' were described by its `different states'. That measurements would intrinsically play a role, also in the description of the reality of a physical entity, only became clear in quantum physics for the case of micro-physical entities.  

In psychology, things historically evolved in a different way. Here, one is in fact confronted with what we call `conceptual entities', such as `concepts' or `conceptual combinations', and more generally with `any cognitive situation which is presented to the different participants in a psychology experiment. Due to their nature, conceptual entities and cognitive situations are `much less real than physical entities', which makes the notion of `state of a conceptual  entity' a highly non-obvious one in psychology. And, as far as we know, the notion of state is never explicitly introduced in psychology, although it appears implicitly within the reasoning that is made about experiments, their setups and results.
Possibly, the notion of `preparation of the experiment' will be used for what we call `the state of the considered conceptual entity' in our approach. Often, however, the notion of state is also associated with the `belief system' of the participant in the experiment. In our approach we keep both notions of
`state' and `measurement' on equal footing, whether our description concerns a physical entity or a conceptual entity. In this way, we can make optimal use of the characteristic methodological strengths of each one of the notions.  It is in doing so that we observed that there is an impressive analogy between the operational-realistic description of a physical entity and the operational-realistic description of a conceptual entity, in particular for what concerns the measurement process and the effects of context on the state of the entity.  As a matter of fact, one can give a SCoP description of a conceptual entity and its dynamics \cite{ga2002,ag2005a,ag2005b}. This justifies the investigation of quantum theory as a unified, coherent and general framework to model conceptual entities, as quantum theory is a natural candidate to model context effects and context-induced state transformations. Hence, the quantum theoretical models that we worked out for specific cognitive situations strictly derive from such investigation of quantum theory as a scientific paradigm for human cognition. In this respect, we think that each predictive success of quantum modeling can be considered as a confirmation of such general validity. It is however important to observe that, recently, potential deviations from Hilbert space modeling were discovered in two cognitive situations, namely, `question order effects' \cite{pnas} and `response replicability' \cite{plosone}. According to some authors, question order effects can be represented by sequential quantum measurements of incompatible properties \cite{bb2012,pb2013,pnas}. However, such a representation seems to be problematical, as it cannot reproduce the pattern observed in response replicability \cite{plosone}, nor it can exactly fit experimental data \cite{bketal2015, asdb2015a}. We put forward an alternative solution for these effects within a `hidden measurement formalism' elaborated by ourselves (see, e.g., \cite{a86, asdb2014,asdb2015b,asdb2015c,a1998, a1999} and references therein), which goes beyond the Hilbert space formulation of quantum theory (probabilities), though it remains compatible with our operational-realistic description of conceptual entities \cite{asdb2015a,asdb2015d}.

For the sake of completeness, we summarize the content of this paper in the following.

In Section~\ref{foundations}, we present the epistemological foundations of the quantum theoretical approach to human cognition we developed in Brussels. We operationally describe a conceptual entity in terms of concrete experiments that are performed in psychological laboratories. Specifically,  the conceptual entity is the reality of the situation which every participant in an experiment is confronted with, and the different states of this conceptual entity are the different modes of reality of this experimental situation. There are contexts influencing the reality of this experimental situation, and the relevant ones of these contexts are elements of the SCoP structure, the theory of our approach, and their influence on the experimental situation is described as a change of state of the conceptual entity under consideration. There are also properties of this experimental situation, the relevant ones being elements of the SCoP structure, and they can be actual or potential, their `amount of actuality' (i.e. their `degree of availability in being actualized') being described by a probability measure.  The operational analogies between physical and conceptual entities suggest to represent the latter by means of the mathematical formalism of quantum theory in Hilbert space. Hence, we assume, in our research, the validity of quantum theory as a scientific paradigm for human cognition. On the basis of this assumption, we provide a unified presentation in Section \ref{Hilbertmodeling} of the results obtained within a quantum theoretical modeling in knowledge representation, decision theory under uncertainty and behavioral economics. We emphasize that our research allowed us to identify new unexpected deviations from classical structures \cite{asv2015,PhilTransA2015}, as well as new genuine quantum structures in conceptual combinations \cite{as2014,asSpinWind2015,IQSA1}, which could not have been identified 
at the same fundamental level as it was possible in our approach
if we would have adopted the 
more traditional perspective only inquiring into the observed deviations from classical probabilistic structures. In Section~\ref{challenges}, we analyze question order effects and response replicability and explain why a quantum theoretical modeling in Hilbert space of these situations is problematical. Finally, we present in Section~\ref{beyond} a novel solution we recently elaborated for these cognitive situations \cite{asdb2015a,asdb2015d}. The solution predicts a violation of the Hilbert space formalism, more specifically, the Born rule for probabilities is put at stake. We however emphasize that this solution remains  compatible with the general operational and realistic description of cognitive entities and their dynamics  given in Section~\ref{foundations}.  In Section~\ref{final}, we conclude our article by offering a few additional remarks, further emphasizing the coherence and advantage of our theoretical approach.  We stress, to conclude this section, that the deviation above from Hilbert space modeling should not be considered as an indication that we should better come back to more traditional classical approaches. On the contrary, we believe that new mathematical structures, more general than both pure classical and pure quantum structures, will be 
needed in the modeling of cognitive processes.

\section{An operational-realistic foundation of cognitive psychology\label{foundations}}
Many quantum physicists agree that the phenomenology of microscopic particles is intriguing, but what is equally curious is the quantum mathematics that captures the mysterious quantum phenomena. Since the early days of quantum theory, indeed, scholars have been amazed by the the success of the mathematical formalism of quantum theory, as it was not clear at all how it had come about. This has inspired a long-standing research on the foundations of the Hilbert space formalism of quantum theory from physically justified axioms, resting on well defined empirical notions, more directly connected with the operations that are usually performed in a laboratory. Such an operational justification would make the formalism of quantum theory more firmly founded.

One of the well-known approaches to the foundations of quantum physics and quantum probability is the `Geneva-Brussels approach', initiated by Jauch \cite{j68} and Piron \cite{p76}, and further developed by our Brussels research team (see, e.g., \cite{a82,a99a}). This research produced a formal approach, called `State Context Property' (SCoP) formalism, where any physical entity can be expressed in terms of the basic notions of `state', `context' and `property', which arise as a consequence of concrete physical operations on macroscopic apparatuses, such as preparation and registration devices, performed in spatio-temporal domains, such as physical laboratories. Measurements, state transformations, outcomes of measurements, and probabilities can then be expressed in terms of these more fundamental notions. If suitable axioms are imposed on the mathematical structures underlying the SCoP formalism, then the Hilbert space structure of quantum theory emerges as a unique mathematical representation, up to isomorphisms \cite{bc81}. 

There are still difficulties connected with the interpretation of some of these axioms and their physical justification, in particular for what concerns compound physical entities \cite{a82}. But, 
but this research line was a source of inspiration for the operational approaches applying the quantum formalism outside the microscopic domain of quantum physics \cite{aa1995,aabg2000}. In particular, 
as we already mentioned in the Introduction,
a very similar realistic and operational representation of conceptual entities can be given for the cognitive domain, in the sense that the SCoP formalism can again be employed to formalize the more abstract conceptual entities in terms of states, contexts, properties, measurements and probabilities of outcomes \cite{ga2002,ag2005a,ag2005b}.

Let us first consider the empirical phenomenology of cognitive psychology. Like in physics, where laboratories define precise spatio-temporal 
domains, we can introduce `psychological laboratories' where cognitive experiments are performed. These experiments are performed on 
situations that are specifically `prepared' for the experiments, including experimental devices, and, for example, structured questionnaires, human participants that interact with the questionnaires in written answers, or each other, e.g., an interviewer and an interviewed. Whenever empirical data are collected from the responses of several participants, a statistics of the obtained outcomes arises. Starting from these empirical facts, we identify in our approach entities, states, contexts, measurements, outcomes and probabilities of outcomes, as follows. 

The complex of experimental procedures conceived by the experimenter, the experimental design and setting and the cognitive effect that one wants to analyze, define a conceptual entity $A$, and are usually associated with a preparation procedure of a state of $A$. Hence, like in physics, the preparation procedure sets the 
initial
state $p_A$ of the conceptual entity $A$ under study. Let us consider, for example, a questionnaire where a participant is asked to rank on a 7-point scale the membership of a list of items with respect to the concepts {\it Fruits}, {\it Vegetables} and their conjunction {\it Fruits and Vegetables}. The questionnaire defines the states $p_{Fruits}$, $p_{Vegetables}$ and $p_{Fruits \ and \ Vegetables}$ of the conceptual entities {\it Fruits}, {\it Vegetables} and {\it Fruits and Vegetables}, respectively. It is true that cognitive situations exist where the preparation procedure of the state of a conceptual entity is hardly controllable. Notwithstanding this, the state of the conceptual entity, defined by means of such a preparation procedure, is a `state of affairs'. It indeed expresses a `reality of the conceptual entity', in the sense that, once prepared in a given state, such condition is independent of any measurement procedure, and can be confronted with the different participants in an experiment, leading to outcome data and their statistics, exactly like in physics.

A context $e$ is an element that can provoke a change of state of the conceptual entity. For example, the concept {\it Juicy} can function as a context for the conceptual entity {\it Fruits} leading to {\it Juicy Fruits}, which can then be considered as a state of the conceptual entity {\it Fruits}. A special context is the one introduced by the measurement itself. Indeed,  when the cognitive experiment starts, an interaction of a cognitive nature occurs between the conceptual entity $A$ under study and a participant in the experiment, in which the state $p_{A}$ of the conceptual entity $A$ generally changes, being transformed to  another state $p$. Also this cognitive interaction is formalized by means of a context $e$. For example, if the participant is asked to choose among a list of items, say, {\it Olive}, {\it Almond}, {\it Apple}, etc., the most typical one with respect to {\it Fruits}, and the answer is {\it Apple}, then the
initial
state $p_{Fruits}$ of the conceptual entity {\it Fruits} changes to $p_{Apple}$, i.e. the state describing the situation `the fruit is an apple',  as a consequence of the contextual interaction with the participant.

The change of the state of a conceptual entity due to a context may be either `deterministic', hence in principle predictable under the assumption that the state before the context acts is known, or `intrinsically probabilistic', in the sense that only the probability $\mu(p,e,p_{A})$ that the  state $p_{A}$ of $A$ changes to the state $p$ is given. In the example above on typicality estimations, the typicality of the item {\it Apple} for the concept {\it Fruits} is formalized by means of the transition probability $\mu(p_{Apple},e, p_{Fruits})$, where the context $e$ is the context of the typicality measurement.

Like in physics, an important role is played by experiments with only two outcomes, the so-called `yes-no experiments'. Suppose that in an opinion poll a participant is asked to answer the question: ``Is Gore honest and trustworthy?''. Only two answers are possible: `yes' and `no'. Suppose that, for a given participant, the answer is `yes'. Then, the state  $p_{Honesty}$ of the conceptual entity {\it Honesty and Trustworthiness} (which we will  
denote by {\it Honesty}, for the sake of simplicity) changes to a new state 
$p_{Gy}$, which is the state describing the situation `Gore is honest'.
Hence, we 
can
distinguish a class of  yes-no measurements on conceptual entities, as we do in physics. 

The third step is the mathematical representation. We have seen that the Hilbert space formalism of quantum theory is general enough to capture an operational description of any entity in the micro-physical domain. Then, the strong analogies between the realistic and operational descriptions of physical and conceptual entities, in particular for what concerns the measurement process, suggest us to apply the same Hilbert space formalism when representing cognitive situations. Hence, each conceptual entity $A$ is associated with a Hilbert space ${\mathcal H}$, and the state $p_A$ of $A$ is represented by a unit vector $|A\rangle \in \mathcal H$. A yes-no measurement is represented by a spectral family $\{M, \mathbbmss{1}-M \}$, where $M$ denotes an orthogonal projection operator over the Hilbert space $\mathcal H$, and $\mathbbmss{1}$ denotes the identity operator over $\mathcal H$. The probability that the `yes' outcome is obtained in such a yes-no measurement when the conceptual entity $A$ is in the state represented by $|A\rangle$ is 
then
given by the Born rule $\mu(A)=\langle A|M|A \rangle$. For example, $M$ may represent an item $x$ that can be chosen in relation to a given concept $A$, so that its membership weight is given by $\mu(A)$. 

The Born rule obviously applies to measurement with more than two outcomes too.  For example, a typicality measurement involving a list of  $n$ different items $x_1$, \ldots, $x_n$ with respect to a concept $A$ can be represented as a spectral measure $\{M_1, \ldots, M_n \}$, where $\sum_{k=1}^n M_k = \mathbbmss{1}$ and $M_kM_l = \delta_{kl}M_k$, such that the typicality $\mu_{k}(A)$ of the item $x_k$ with respect to the concept $A$ is again given by the Born rule $\mu_{k}(A)=\langle A|M_k|A\rangle$.

An interesting aspect concerns the final state of a conceptual entity $A$ after a human judgment. As above, we can assume the existence of a nonempty class of cognitive measurements that are ideal first kind measurements in the standard quantum sense, i.e. that satisfy the `L\"{u}ders postulate'. For example, if the typicality measurement of a list of items $x_1$, \ldots, $x_n$ with respect to a concept $A$ gave the outcome $x_k$, then the final state of the conceptual entity after the measurement is represented by the unit vector $|A_k\rangle=\frac{M_{k}|A\rangle}{\sqrt{\langle A |M_{k}|A\rangle}}$. This means that the weights $\mu_{k}(A)$ given by the Born rule can actually be interpreted as transition probabilities $\mu(p_k,e,p_{A})$, where $e$ is the context producing the transitions from the initial state $p_{A}$ of the conceptual entity $A$, represented by the unit vector $|A\rangle$, to one of the $n$ possible outcome states $p_k$, represented by the unit vectors $|A_k\rangle$.
 
So, how can a Hilbert space model be actually constructed for a cognitive situation? To answer this question let us consider  again a conceptual entity $A$, in the state $p_{A}$, a cognitive measurement on $A$ described by means of  a context $e$, and suppose that the measurement has $n$ distinct outcomes, $x_1$, $x_2$, \ldots, $x_n$. A quantum theoretical model for this situation can be constructed as follows. We associate $A$ with a
$n$-dimensional complex Hilbert space $\cal H$, and then consider an orthonormal base $\{ |e_1\rangle, |e_2\rangle, \ldots, |e_n\rangle \}$ in $\cal H$ (since ${\cal H}$ is isomorphic to the Hilbert space ${\mathbb C}^{n}$, the orthonormal base of $\cal H$ can be the canonical base of ${\mathbb C}^{n}$). Next, we represent the cognitive measurement described by $e$ by means of the spectral family $\{M_1, M_2, \ldots, M_n \}$, where $M_k=|e_k\rangle\langle e_k|$, $k=1,2,\ldots,n$. Finally, the probability that the measurement $e$ on the conceptual entity $A$ in the state $p_{A}$
gives the outcome $x_k$ is given by $\mu_{k}(A)=\langle A|M_k|A\rangle=\langle A|e_k\rangle\langle e_k|A\rangle=|\langle e_k|A\rangle|^2$.

What about the interpretation of the Hilbert space formalism above? Two major points should now be reminded, namely: 

(i) the states of conceptual entities describe the `modes of being' of these conceptual entities;

(ii) in a cognitive experiment, a participant 
acts as a (measurement) context for the conceptual entity, changing its state.

This means that, as we mentioned already, the  state $p_{A}$ of the conceptual entity $A$ is represented in the Hilbert space formalism by the unit vector $|A\rangle$, the possible outcomes $x_k$ of the experiment by the base vectors $|e_k\rangle$, and the action of a participant (or the overall action of the ensemble of participants) as the state transformation $|A\rangle \to |e_k\rangle$ induced by the orthogonal projection operator $M_k=|e_k\rangle\langle e_k|$, if the outcome $x_k$ is obtained, so that the probability of occurrence of $x_k$ can also be written as $\mu_{k}(A)=\mu(|e_k\rangle,e,|A\rangle)$, where $e$ is the measurement context associated with the spectral family $\{M_1, M_2, \ldots, M_n \}$.

It follows from (i) and (ii) that a state, hence a unit vector in the Hilbert space representation of states, does not describe the subjective 
beliefs 
of a person, or collection of persons, about a conceptual entity. Such subjective  beliefs are rather incorporated in the cognitive interaction between the cognitive situation and the human participants deciding on that cognitive situation. In this respect, our operational quantum approach to human cognition  is also a realistic one, and thus it departs from other approaches that apply the mathematical formalism of quantum theory to model cognitive processes \cite{k2010,bb2012,hk2013,pb2013,pnas,plosone}. Of course, one could say that the difference between interpreting the quantum state as a `state of belief' of a participant in the experiment, or as a `state of a conceptual entity', i.e. a `state of the situation which the participant is confronted with during an experiment', is only a question of philosophical interpretation, but comes to the same when it concerns the methodological development of the approach. Although this is definitely partly true, we do not fully agree with it. Interpretation and methodology are never completely separated. A certain interpretation, hence giving rise to a specific view on the matter, will give rise to other ideas of how to further develop the approach, how to elaborate the method, etc., than another interpretation, with another view, will do. We believe that an operational-realistic approach, being balanced between attention for idealist as well as realist philosophical 
interpretations,
carries in this sense a particular strength, 
precisely
due to this balance. A good example of this is how we were inspired to use the superposition principle of quantum theory in our modeling of concepts as conceptual entities. We represented the combination of two concepts by a state that is the linear superposition of the states describing the component concepts. This way of representing combined conceptual entities captures the nature of emergence, exactly like in physics. It would not be obvious to put forward 
this
description when state of beliefs are the focus of what can be predicted.

We stress a third point that is important, in our opinion. For most situations, we interpret the effect of the cognitive context on a conceptual entity in a decision-making process as an `actualization of pure potentiality'. Like in quantum physics, the (measurement) context does not reveal pre-existing properties of the entity but, rather, it makes actual properties that were only potential in the initial state of the entity (unless the initial state is already an eigenstate of the measurement in question, like in physics) \cite{ga2002,ag2005a,ag2005b}.

It follows from the previous discussion that our research investigates the validity of quantum theory as a general, unitary and coherent theory for human cognition. Our quantum theoretical models, elaborated for specific cognitive situations and data, derive from quantum theory as a consequence of the assumptions about this general validity. As such, these models are subject to the technical and epistemological constraints of quantum theory. 
In other terms,
our quantum modeling rests on a `theory based approach', and should be distinguished from an `ad hoc modeling based approach', only devised to fit data. In this respect, one should be suspicious of models in which free parameters are added on an `ad hoc' basis to fit the data more closely in specific experimental situations. In our opinion, the fact that our `theory derived model' reproduces different sets of experimental data constitutes in itself a convincing argument to support its advantage over traditional modeling approaches and to extend its use to more complex cognitive situations (in that respect, see also our final remarks in Section~\ref{final}).

We present in Section \ref{Hilbertmodeling} the results obtained in our quantum theoretical approach in the light of the epistemological perspective of this section.

\section{On the modeling effectiveness of Hilbert space\label{Hilbertmodeling}}
The quantum approach to cognition described  in Section \ref{foundations} produced concrete models in Hilbert space, which faithfully matched different sets of experimental data collected to reveal `decision-making errors' and `probability judgment errors'. This allowed us to identify genuine quantum structures in the cognitive realm. We present a reconstruction of the attained results in the following.

The first set of results concerns knowledge representation and conceptual categorization and combination. James Hampton collected data on how people rate membership of items with respect to pairs of concepts and their combinations, conjunction \cite{h1988a}, disjunction \cite{h1988b} and negation \cite{h1997}. By using the data in \cite{h1988b}, we reconstructed the typicality estimations of 24 items with respect to the concepts {\it Fruits} and {\it Vegetables} and their disjunction {\it Fruits or Vegetables}. We showed that the concepts {\it Fruits} and {\it Vegetables} interfere when they combine to form {\it Fruits or Vegetables}, and the state of the latter can be represented by the linear superposition of the states of the former.  This behavior is analogous to that of quantum particles interfering in the double-slit experiment when both slits are open. The data are faithfully represented in a 25-dimensional Hilbert space over complex numbers \cite{abgs2013,ags2013}.

In the data collected on the membership estimations of items with respect to pairs $(A,B)$ of concepts and their conjunction `$A$ and $B$' and disjunction `$A$ or $B$', Hampton found systematic violations of the rules of classical (fuzzy set) logic and probability theory. For example, the membership weight of the item {\it Mint} with respect to the conjunction {\it Food and Plant} is higher than the membership weight of {\it Mint} with respect to both {\it Food} and {\it Plant} (`overextension'). Similarly, the membership weight of the item {\it Ashtray} with respect to the disjunction {\it Home Furnishing or Furniture} is lower than the membership weight of {\it Ashtray} with respect to both {\it Home Furnishing} and {\it Furniture} (`underextension'). We showed that overextension and underextension are natural expressions of `conceptual emergence' \cite{a2009a,ags2013}. Namely, whenever a person estimates the membership of an item  $x$ with respect to the pair $(A,B)$ of concepts and their combination $C(A,B)$, two processes act in the person's mind. The first process is guided by `emergence', that is, the person estimates the membership of  $x$ with respect to the new emergent concept $C(A,B)$. The second process is guided by `logic', that is, the person separately estimates the membership of $x$ with respect to $A$ and $B$ and applies 
a probabilistic logical calculus to estimate the membership of $x$ with respect to $C(A,B)$  \cite{IQSA2}. More important, the new concept $C(A,B)$ emerges from the concepts $A$ and $B$, exactly as the linear superposition of two quantum states emerges from the component states. A two-sector Fock space faithfully models Hampton's data, and was later successfully applied to the modeling of more complex situations involving concept combinations (see, e.g., \cite{IQSA2,s2015}).

It is interesting to note that the size of deviation of classical probabilistic rules due to overextension and underextension generally depends on the item  $x$ 
and the specific combination $C(A,B)$ of the concepts $A$ and $B$ that are investigated. However, we recently performed a more general experiment in which we asked the participants to rank the membership of items with respect to the concepts $A$, $B$, their negations `not $A$', `not $B$', and the conjunctions `$A$ and $B$', `$A$ and not $B$', `not $A$ and $B$', and `not $A$ and not $B$'.  We surprisingly found that the size of deviation from classicality in this experiment does not depend on either the item or the pair of concepts or the specific combination, but shows to be a numerical constant. Even more surprisingly, our two-sector Fock space model correctly predicts the value of this constant, capturing in this way a deep non-classical mechanism connected in a fundamental way with the mechanism of conceptual formation itself rather than only specifically with the mechanism of conceptual combination \cite{asv2015,PhilTransA2015}.

Different concepts entangle when they combine, where `entanglement' is meant in the standard quantum sense. We proved this feature of concepts in two experiments. In the first experiment, we asked the participants to choose the best example for the conceptual combination {\it The Animal Acts} in a list of four examples, e.g., {\it The Horse Growls}, {\it The Bear Whinnies}, {\it The Horse Whinnies} and {\it The Bear Growls}. By suitably combining exemplars of {\it Animal} and exemplars of {\it Acts}, we performed four joint measurements on the combination {\it The Animal Acts}. The expectation values violated the `Clauser-Horne-Shimony-Holt' version of Bell inequalities \cite{b1964,chsh1969}. The violation was such that, not only the state of {\it The Animal Acts} was entangled, but also the four joint measurements were entangled, in the sense that they could not be represented in the Hilbert space ${\mathbb C}^{4}$ as the (tensor) product of a measurement performed on the concept {\it Animal} and a measurement performed on the concept {\it Acts} \cite{as2014}. In the second experiment, performed on the conceptual combination {\it Two Different Wind Directions}, we confirmed the presence of quantum entanglement, but we were also able to prove that the empirical violation of the marginal law in this type of experiments is due to a bias of the participants in picking wind directions. If this bias is removed, which is what we did in an ensuing experiment on {\it Two Different Space Directions}, one can show that people pick amongst different space directions exactly as coincidence spin measurement apparatuses pick amongst different spin directions of a compound system in the singlet spin state. In other words, entanglement in concepts can be proved from only the statistics of the correlations of joint measurements on combined concepts, exactly as in quantum physics \cite{asSpinWind2015}.

Since concepts exhibit genuine quantum features when they combine pairwise, it is reasonable to expect that these features should be reflected in the statistical behavior of the combination of several identical concepts. Indeed, we detected quantum-type indistinguishability in an experiment on the combination of identical concepts, such as the combination {\it Eleven Animals}. More specifically, we found significant evidence of deviation from the predictions of classical statistical theories, i.e. `Maxwell-Boltzmann distribution'. This deviation has 
clear analogies with the deviation of quantum mechanical from classical mechanical statistics, due to indistinguishability of microscopic quantum particles, that is, we found convincing evidence of the presence of `Bose-Einstein distribution'. In the experiment, indeed, people do 
not seem to distinguish two identical concepts in the combination of $N$ identical concepts, which is more evident in more abstract than in more concrete concepts, as expected \cite{IQSA1}. 

The second set of results concern `decision-making errors under uncertainty'. In the `disjunction effect' people prefer action $x$ over action $y$ if they know that an event $A$ occurs, and also if they know that $A$ does not occur, but they prefer $y$ over $x$ if they do not know whether $A$ occurs or not. The disjunction effect violates a fundamental principle of rational decision theory, Savage's `Sure-Thing principle' and, more generally, the total probability rule of classical probability \cite{s1954}. This preference of sure over unsure choices violating the Sure-Thing principle was experimentally detected in the `two-stage gamble' and in the `Hawaii problem' \cite{ts1992}. In the experiment on a gamble that can be played twice, the majority of participants prefer to bet again when they know they won in the first gamble, and also when they know they lost in the first gamble, but they generally prefer not to play when they do not know whether they won or lost. In the Hawaii problem, most students decide to buy the vacation package when they know they passed the exam, and also when they know they did not pass the exam, but they generally decide not to buy the vacation package when they do not know whether they passed or not passed the exam. We recently showed that, in both experimental situations, this `uncertainty aversion' can be explained as an effect of underextension of the conceptual entities $A$ and `not $A$' with respect to the conceptual disjunction `$A$ or not $A$', where the latter describes the situation of not knowing which event, $A$ or `not $A$', will occur. The concepts $A$ and `not $A$' interfere in the disjunction `$A$ or not $A$', which determines its underextension. A Hilbert space model in ${\mathbb C}^{3}$ allowed us to reproduce the data in both experiments on the disjunction effect \cite{s2015}.

Ellsberg's thought experiments, much before the disjunction effect, revealed that the Sure-Thing principle is violated in concrete decision-making under uncertainty, as people generally prefer known over unknown probabilities, instead of maximizing their expected utilities. In the famous `Ellsberg three-color example', an urn contains 30 red balls and 60 balls that are either yellow or black, in unknown proportion. One ball will be drawn at random from the urn. The participant is firstly asked to choose between betting on `red' and betting on `black'. Then, the same participant is asked to choose between betting on `red or yellow' and betting on `black or yellow'. In each case, the `right' choice will be awarded with $\$100$. As the events `betting on red' and `betting on black or yellow' are associated with known probabilities, while their counterparts are not, the participants will prefer betting on the former than betting on the latter, thus revealing what Ellsberg called `ambiguity aversion', and violating the Sure-Thing principle \cite{e1961}. This pattern of choice has been confirmed by several experiments in the last thirty years \cite{ms2014}. Recently, Machina identified in a couple of thought experiments, the `50/51 example' and the `reflection example', a similar mechanism guiding human preferences in specific ambiguous situations, namely, `information symmetry' \cite{m2009,blhp2011}, which was experimentally confirmed in \cite{lhp2010}. In our quantum theoretical approach, ambiguity aversion and information symmetry are two possible cognitive contexts influencing human preferences in uncertainty situations and changing the states of the `Ellsberg and Machina conceptual entities', respectively. Hence, an ambiguity aversion context will change the state of the Ellsberg conceptual entity in such a way that `betting on red' and `betting on black or yellow' are finally preferred. In other terms, the novel element of this approach is that the initial state of the conceptual entity, in its Hilbert space representation, can also change because of the pondering of the participants in relation to certain choices, before being collapsed into a given outcome.  This opens the way to a generalization of rational decision theory with quantum, rather than classical, probabilities \cite{asJMP2015}.

The results above provide a strong confirmation of 
the quantum theoretical approach presented in Section \ref{foundations}, and we expect that further evidence will be given in this direction in the years to come. In the next section we instead intend to analyze some situations where deviations from Hilbert space modeling of human cognition apparently occur.
We will see in Section~\ref{beyond} that these deviations are however compatible with the general  operational-realistic framework  portrayed in Section~\ref{foundations}.

\section{Deviating from Hilbert space\label{challenges}} 
As mentioned in Section~\ref{foundations}, if suitable axioms are imposed on the SCoP formalism, the Hilbert space structure of quantum theory can be shown to emerge uniquely, up to isomorphisms \cite{bc81}. However, we also know that certain experimental situations can violate some of these axioms. This is the case for instance when we consider entities formed by experimentally separated sub-entities, a situation that cannot be described by the standard quantum formalism \cite{a82,a1984}. Similarly, one may expect that the structural shortcomings of the standard quantum formalism can also manifest in the ambit of psychological measurements, in the form of data that cannot be exactly modeled (or jointly modeled) by means of the specific Hilbert space geometry and the associated Born rule. The purpose of this section is to describe two paradigmatic examples of situations of this kind: `question order effects' and `response replicability'. In the following section, we then show how the quantum formalism can be naturally completed to also faithfully model these data, in a way that remains consistent with our operational-realistic approach. 

Let us first remark that the mere situation of having to deal with a set of data for which we don't have yet a faithful Hilbert space model should not make one necessarily search for an alternative more general quantum-like mathematical structure as a modeling environment. Indeed, it is very well possible that the adequate Hilbert space model has not yet been found. Recently, however, a specific situation was identified and analysed indicating that the standard quantum formalism in Hilbert space would not be able to be used to model it \cite{plosone}. This situation combines two phenomena: `question order effects' and `response replicability'. We start by explaining `question order effects' and how the cognitive situation in which they appear can be represented in Hilbert space.

For this we come back to the yes-no experiment of Section~\ref{foundations}, where participants are asked: ``Is Gore honest and trustworthy?''. This experiment gives rise to a two-outcome measurement performed on the conceptual entity {\it Honesty} in the initial state $p_H$, represented by the unit vector $|H\rangle\in {\mathcal H}$, where ${\mathcal H}$ is a two-dimensional Hilbert space if we assume the measurement to be non-degenerate, or more generally a $n$-dimensional Hilbert space if we also admit the possibility of sub-measurements. Denoting $\{M_G, \bar{M}_G = \mathbbmss{1}-M_G\}$ the spectral family associated with this measurement, the probability of the `yes' outcome (i.e. to answer `yes' to the question about Gore's honesty and trustworthiness) is then given by the Born rule $\mu_{Gy}(H)=\langle H|M_G|H\rangle$, and of course $\mu_{Gn}(H)=\langle H|\bar{M}_G|H\rangle = 1-\mu_{Gy}(H)$ is the probability for the `no' outcome. We then consider a second measurement performed on the conceptual entity {\it Honesty}, but this time associated with the question: ``Is Clinton honest and trustworthy?''. We denote $\{M_C, \bar{M}_C = \mathbbmss{1}-M_C\}$ the spectral family associated with this second measurement, so that the probabilities for the `yes' and `no' outcomes are again given by $\mu_{Cy}(H)=\langle H|M_C|H\rangle$ and $\mu_{Cn}(H)=\langle H|\bar{M}_C|H\rangle$, respectively. 

Starting from these two measurements, it is possible to conceive sequential measurements, corresponding to situations where the respondents are subject to the Gore and Clinton questions in a succession, one after the other, in different orders. Statistical data about `Clinton/Gore' sequential measurements were reported in a seminal article on question order effects \cite{m2002} and further analyzed in \cite{bb2012,wb2013}. More precisely, after fixing a rounding error in \cite{wb2013}, we have the following sequential (or conditional) probabilities \cite{asdb2015a}: 
\begin{eqnarray}
&\mu_{CyGy}(H)=0.4899 \quad \mu_{CyGn}(H)= 0.0447 \quad \mu_{CnGy}(H)=0.1767 \quad \mu_{CnGn}(H)= 0.2887 \label{Prob-Clinton-Gore1}\\
&\mu_{\rm GyCy}(H)=0.5625 \quad \mu_{GyCn}(H) = 0.1991 \quad \mu_{GnCy}(H) = 0.0255 \quad \mu_{GnCn}(H) = 0.2129
\label{Prob-Clinton-Gore2}
\end{eqnarray}
where (\ref{Prob-Clinton-Gore1}) corresponds to the sequence where first the Clinton and then the Gore measurements are performed, whereas (\ref{Prob-Clinton-Gore2}) corresponds to the reversed order sequence for the measurements. Considering that the probabilities in each of the four columns above are sensibly different, these data describe typical `question order effects'. 

Quantum theory is equipped with a very natural tool to model question order effects: `incompatible measurements', as expressed by the fact that two self-adjoint operators, and the associated spectral families, in general do not commute. More precisely, the Hilbert space expression for the probability that, say, we obtain the answer $CyGn$ when we perform first the Clinton measurement and then the Gore one, is \cite{bb2012,wb2013}: $\mu_{CyGn}(H)= \langle H|M_C\bar{M}_GM_C|H\rangle$. Similarly, the probability to obtain the outcome $GnCy$, for the sequential measurement in reversed order, is: $\mu_{GnCy}(H)= \langle H|\bar{M}_GM_C\bar{M}_G|H\rangle$. Since we have the operatorial identity $\bar{M}_GM_C\bar{M}_G-M_C\bar{M}_GM_C=(M_G-M_C)[M_G,M_C]$, the difference $\mu_{GnCy}(H)-\mu_{CyGn}(H)$ will generally be non-zero if $[M_G,M_C]\neq 0$, i.e. if the spectral families associated with the two measurements do not commute. In the following we will analyse whether non-compatibility within a standard quantum approach can cope in a satisfying way with these question order effects, and show that a simple `yes' to this question is not possible. Indeed, a deep problem already comes to the surface in relation to the phenomenon of `response replicability'.

Consider again the Gore/Clinton measurements: if a respondent says `yes' to the Gore question, then is asked the Clinton question, then again is asked the Gore question, the answer given to the latter will almost certainly be a `yes', independently of the answer given to the intermediary Clinton question. This phenomenon is called `response replicability'. If in addition to question order effects also response replicability is jointly modeled in  Hilbert space quantum mechanics, a contradiction can be detected, as shown in \cite{plosone}. Let us indicate what are the elements that produce this contradiction. In standard quantum mechanics only if a state is an eigenstate of the considered measurement the outcome `yes' will be certain in advance. Also, measurements that can transform an arbitrary initial state into an eigenstate are ideal measurements called of the first kind.  According to response replicability, outcomes that once have been obtained for a measurement will have to become certain in advance if this same measurement is performed a second time. This means that the associated measurements should be ideal and of the first kind. For the case of the Gore/Clinton measurements, and the situation of response replicability mentioned above, this means that the Gore measurement should be ideal and of the first kind. But  one can also consider the situation 
where first the Clinton measurement is performed, then the Gore measurement and afterwards the Clinton measurement again. A similar analysis leads then to the Clinton measurement needing to be ideal and of the first kind. This means however that after more than three measurements that alternate between Clinton and Gore, the state needs to have become an eigenstate of both measurements. As a consequence, both measurements can be shown to be represented by commuting operators. The proof of the contradiction between `response replicability' and 'non-commutativity' worked out in \cite{plosone} is formal and also more general than the intuitive reasoning presented above -- for example, the contradiction is also proven when measurements are represented by positive-operator valued measures instead of projection valued measures, which is what we have considered here -- and hence indicates that the non-commutativity of the self-adjoint operators needed to account for the question order effects cannot be realised together with the `ideal and first kind' properties
needed to account for the response replicability within a standard quantum Hilbert space setting.      

Although refined experiments would be needed to reveal the possible reasons for response replicability, it is worth to put forward some intuitive ideas, as we have been developing a quantum-like but more general than Hilbert space formalism within our Brussels approach to quantum cognition \cite{asdb2014,asdb2015b,asdb2015c}, and we believe that we can cope with the above contradiction within this more general quantum-like setting in a very natural way. It seems to be a plausible hypothesis that response replicability is, at least partly, due to a multiplicity of effects, that however take place during the experiment itself, such as desire of coherence, learning, fear of being judged when changing opinion, etc. And a crucial aspect for both  question order effects and response replicability appearing
in the Gore/Clinton situation is that the sequential measurements need to be carried out with the same participant, who has to be tested again and again. This is different than
the situation in quantum physics, where order effects appear for non-commuting observables also when sequential measurements are performed with different apparatuses. 
Hence, both question order effects and response replicability seem to be the consequence of `changes taking place in the way each subject responds probabilistically to the situation -- described by the state of the conceptual entity in our approach -- he or she is confronted with during a measurement'. Since the structure of the probabilistic response to a specific state is fixed in quantum mechanics, being determined by the Born rule, it is clear that such a change of the probabilistic response to a given measurement, when it is repeated in a sequence of measurements, cannot be accounted for by the standard quantum formalism. And it is exactly such structure of the probabilistic response to a same measurement with respect to a given state that can be varied in the generalized quantum-like theory that we have been developing \cite{asdb2014,asdb2015b,asdb2015c}. This is the reason that, when we became aware of the contradiction identified in \cite{plosone}, we were tempted to investigate whether in our generalized quantum-like theory the contradiction would vanish, and response replicability would be jointly
modelizable with question order effects. And indeed, we could obtain a positive result with respect to this issue \cite{asdb2015a}, which we will now sketch in the next section.

\section{Beyond-quantum models\label{beyond}}
We presented in Section \ref{challenges} two paradigmatic situations in human cognition that cannot be modeled together using the standard quantum formalism. We want now to explain how the latter can be naturally extended to also deal with these situations, still remaining in the ambit of a unitary and coherent framework for cognitive processes. 

For this, we introduce a formalism where the probabilistic response with respect to a specific experimental situation, i.e. a state of the conceptual entity under consideration, can vary, and hence can be different than the one compatible with the Born rule of standard quantum theory. This formalism, called 
the `extended Bloch representation' of quantum mechanics \cite{asdb2014}, 
exploits
in its most recent formulation the fact that the states of a quantum entity (described as ray-states or density matrix-states) can be uniquely mapped into a convex portion of a generalized unit Bloch sphere, in which also measurements can be represented in a natural way, by means of appropriate simplexes having the eigenstates as vertex vectors. A measurement can then be described as a process during which an abstract point particle (representing the initial state of the quantum entity) enters into contact with the measurement simplex, which then, as if it was an elastic and disintegrable hyper-membrane, can collapse to one of its vertex points (representing the outcomes states) or to a point of one of its sub-simplexes (in case the measurement would be degenerate). 
 
We do not enter here into the details of this remarkable process, and refer the reader to the detailed descriptions  in \cite{asdb2015a,asdb2014,asdb2015b,asdb2015c}. For our present purposes, it 
will be
sufficient to observe that a measurement simplex, considered as an abstract membrane that can collapse as a result of some uncontrollable environmental fluctuations, can precisely model that aspect of a measurement that in the quantum jargon is called `wave function collapse'. More precisely, when the abstract point particle enters into contact with the `potentiality region' represented by such membrane, it creates some `tension lines' partitioning the latter into different subregions, one for each possible outcome. The collapse of the membrane towards one of the vertex points then depends on which subregion disintegrates first, so that the different outcome probabilities can be expressed as the relative Lebesgue measures of these subregions (the larger a subregion, the higher the associated probability). In other terms, this membrane's mechanism, with the tension lines generated by the abstract point particle, is a mathematical representation of a sort of `weighted symmetry breaking' process. Now, thanks to the remarkable geometry of simplexes, it can be proven that if the membrane is chosen to be uniform, thus having the same probability of disintegrating in any of its points (describing the different possible measurement-interactions), the collapse probabilities are exactly given by the Born rule. In other terms, the latter can be derived, and explained, as being the result of a process of actualization of potential hidden-measurement interactions, so that the extended Bloch representation constitutes a possible solution to the measurement problem.

Thus, when the membrane is uniform, the `way of choosing' an outcome is precisely the `Born way'. However, a uniform membrane is a very special situation, and it is natural to also consider membranes whose points do not all
have the same probability of disintegrating, i.e. membranes whose disintegrative processes are described by non-uniform probability densities $\rho$, which we simply call $\rho$-membranes. Non-uniform $\rho$-membranes can produce outcome probabilities different from the standard quantum ones and give rise to probability models different from the Hilbertian one (even though the state space is a generalized Bloch sphere derived from the Hilbert space geometry\footnote{More general state spaces can also be considered, in what has been called the `general tension-reduction' (GTR) model \cite{asdb2015b,asdb2015c,asdb2015d}.}). But this is exactly what one needs in order to account, in a unified framework, for the situation we encounter when combining the phenomena of `response replicability' and `question order effects', as previously described and analysed in \cite{plosone}.

We thus see that it is possible to naturally complete the quantum formalism to obtain a finer grained description of psychological experiments in which the probabilistic response of a measurement with respect to a state can be different to the one described by the Born rule. Additionally, our generalized quantum-like theory also explains why, despite the fact that individual measurements are possibly associated with different non-Born probabilities, the Born rule nevertheless appears to be a very good approximation to describe numerous experimental situations. This is related to the notion of `universal measurement', firstly introduced by one of us in \cite{a1998} and further analyzed in \cite{asdb2014,asdb2015b,asdb2015c,asdb2014b}. In a nutshell, a universal measurement is a measurement whose probabilities are obtained by averaging over the probabilities of all possible quantum-like measurements sharing a same set of outcomes, in a same state space. In other terms, a universal measurement corresponds to an average over all possible non-uniform $\rho$-membranes, associated with a given measurement simplex. Following a strategy similar to that used in the definition of the `Wiener measure', it is then possible to show that if the state space is Hilbertian (more precisely, a convex set of states inscribed in a generalized Bloch sphere, inherited from a Hilbert space), then the probabilities of a universal measurement are precisely those predicted by the Born rule. 

In \cite{asdb2015a} we could show that the joint situation of question order effects and response replicability for the data collected with respect to the Gore/Clinton measurements, and others, is modelizable
within our generalized quantum theory by introducing non-Born type measurements. However, we  were also able to provide a better modeling of the question order effects data as such. Indeed, using standard Born-probability quantum theory it was only possible to model approximately these data in earlier studies \cite{wb2013}. This is due to the existence of a general algebraic equality about sequential measurements in standard quantum mechanics which is the following \cite{asdb2015a,wb2013,Niestegge2008}:
\begin{equation}
Q\equiv M_GM_CM_G - M_CM_GM_C + \bar{M}_G\bar{M}_C\bar{M}_G - \bar{M}_C\bar{M}_G\bar{M}_C=0
\label{QQ-equality}
\end{equation}
where $\{M_G, \bar{M}_G = \mathbbmss{1}-M_G\}$ and $\{M_C, \bar{M}_C = \mathbbmss{1}-M_C\}$ are the spectral families associated with the Hilbert model of the Gore and Clinton measurements introduced in Section \ref{challenges}.
Taking the average $q=\langle H|Q|H\rangle$, one thus obtains, more specifically: 
\begin{equation}
q\equiv\mu_{GyCy}(H) - \mu_{CyGy}(H) + \mu_{GnCn}(H) - \mu_{CnGn}(H) =0.
\label{QQ-equality2}
\end{equation}
This equality has been called the `QQ-equality', and can be used as a test for the quantumness of the probability model, but only in the sense that a quantum model, necessarily, has to obey it, although the fact that it does so is not a guarantee that the model will be Hilbertian. Inserting the experimental values (\ref{Prob-Clinton-Gore1})-(\ref{Prob-Clinton-Gore2}) into (\ref{QQ-equality2}), one finds $q=0.0032\neq 0$. 
This value is small (being only $0.32\%$ of the maximum value $q$ can take, which is 1), which is the reason that approximate modeling can be obtained within standard quantum mechanics \cite{wb2013}. Note however that (\ref{QQ-equality}) does not depend on the dimension of the Hilbert space considered, which means that even in higher dimensional Hilbert spaces, if degenerate measurements 
are
considered, an exact modeling 
would
still be impossible to obtain. We have reasons to believe that also question order effects, with the QQ-equality standing in the way of an exact modeling of the data, contain an indication for the need to turn to a more general quantum-like theory, such as the one we used to cope with the joint phenomenon of question order effects and response replicability. We 
present some arguments in this regard in the following of this section.

First,  
we
note that in case one would choose a two-dimensional Hilbert space, which is the natural choice when dealing with two-outcome measurements, additional equalities can be written that are this time strongly violated by the data. As an example, consider the quantity \cite{asdb2015a}:
\begin{eqnarray}
q'&\equiv& \mu_{CyGn}(H)\mu_{CnGn}(H) - \mu_{CnGy}(H)\mu_{CyGy}(H)\\
&=& \langle H|M_C\bar{M}_GM_C|H\rangle \langle H|\bar{M}_C\bar{M}_G\bar{M}_C|H\rangle - \langle H|\bar{M}_CM_G\bar{M}_C|H\rangle \langle H|M_CM_GM_C|H\rangle
\label{QQ-equality3}
\end{eqnarray}
If the Hilbert space is two-dimensional, one can write $M_G=|G\rangle\langle G|$, $\bar{M}_G=|\bar{G}\rangle\langle \bar{G}|$, as well as $M_C=|C\rangle\langle C|$, $\bar{M}_C=|\bar{C}\rangle\langle \bar{C}|$. Replacing these expressions into (\ref{QQ-equality3}) one finds, after some easy algebra, that $q'=0$. However, inserting the experimental values (\ref{Prob-Clinton-Gore1})-(\ref{Prob-Clinton-Gore2}) into (\ref{QQ-equality3}), one finds $q'=-0.073\neq 0$, which not only 
is not zero, but also $29.2\%$ of the maximum value that $q'$ can take (which is $0.25$).

Second, let us repeat our intuitive reasoning as to why measurements in the situation of response replicability carry non-Bornian probabilities. Due to the local 
contexts
of the collection of sequential measurements, Gore, Clinton, and then Gore again, the third measurement internally changes into a non-Bornian one, and more specifically a deterministic one for the considered state, since response replicability means that for all subsequent Gore measurements the same outcome is assured. It might well be the case, although an intuitive argument would be more complex 
to give
in this case, that also for the situation of question order effects, precisely because they only appear if a same human mind is sequentially interrogated, non-Bornian probabilities would be required. An even stronger hypothesis, which we plan to investigate in the future, is that most individual human minds, and perhaps even all, would carry in general non-Bornian probabilities, so that the success of standard quantum mechanics and Bornian probabilities would be mainly an effect of averaging over a sufficiently large set of different human minds, which effectively is what happens in a standard psychological experiment. If this last hypothesis is true, the violation of the Born rule for question order effects and  response replicability would be quite natural, since the same human mind is needed to provoke these effects. Indeed, our analysis in \cite{asdb2015b,asdb2015c} shows that standard quantum probabilities in the modeling of human cognition can be explained by considering that in numerous experimental situations the average over the different participants will be quite close to that of a universal measurement, which as we observed is exactly given by the Born rule. In other terms, even if the probability model of an individual psychological measurement could be non-Hilbertian, it will generally admit a first order approximation, and when the states of the conceptual entity under investigation can be described by means of a Hilbert space structure, this first order approximation will precisely correspond to the quantum mechanical Born rule. 

If the above considerations provide an interesting piece of explanation as to why the Born rule is generally successful also beyond the micro-physical domain, at the same time it also contains a plausible reason of why it will possibly be not successful in all experimental situations, i.e. when the average is either not large enough, or when the experiment is so conceived that it doesn't apply as such. This could be the typical situation of question order effects and response replicability, since in this case we do not consider an average over single measurements, but over sequential (conditional) measurements. And this could be an explanation of why Hilbertian symmetries like those described above can be easily violated and that it will not be possible, by means of the Born rule, to always obtain an exact fit of the data \cite{asdb2015a,asdb2015d}.

Additionally, as we said, it allowed us to precisely fit the data by using the extended Bloch representation,
and more specifically 
simple one-dimensional locally uniform membranes inscribed in a 3-dimensional Bloch sphere that 
can
disintegrate (i.e. break) only inside a connected internal region \cite{asdb2015a}. Thanks to this modeling, we could also understand that the reason the Clinton/Gore and similar data appear to almost obey the QQ-equality (\ref{QQ-equality2}) is quite different from the reason the equality is obeyed by pure quantum probabilities. Indeed, in a pure quantum model two specific contributions to the $q$-value (\ref{QQ-equality2}), called the `relative indeterminism' and `relative asymmetry' contributions, are necessarily both identically zero, whereas we 
could
show, using our extended model, that for the data (\ref{Prob-Clinton-Gore2}),
and similar data,  
these two contributions are both very different from zero, but happen to almost cancel each other, thus explaining why the $q=0$ equality is almost obeyed, although the probabilities are manifestly non-Bornian \cite{asdb2015a}.

\section{Final considerations\label{final}}
In this article we explained the essence of the operational-realistic approach to cognition developed in Brussels, which in turn originated from the foundational approach to physics elaborated initially in Geneva and then in Brussels (in what has become known as the `Geneva-Brussels school'). Our emphasis was that this approach is sufficiently general, and fundamental, to provide a unitary framework that can be used to coherently describe, and realistically interpret, not only standard quantum theory, but also its natural extensions, like the extended Bloch model and the GTR-model. In this final section we offer some additional comments on our approach to cognition, taking into consideration the confusion that sometimes exists between `ad hoc (phenomenological) models' and `theoretical (first principle) models', as well as the critique that a Hilbertian model (and a fortiori its possible extensions) is suspicious because it allows `too many free parameters' to obtain an exact fit (and not just an approximate fit) for all the experimental data. 

In that respect, it is worth emphasizing that the principal focus of our `theory of human cognition' is not to model as precisely as possible the data gathered in psychological measurements. A faithful modeling of the data is of course an essential part of it, but our aim is actually more ambitious. In putting forward our methodology, consisting in looking at instances of decision-making as resulting from an interaction of a decision-maker with a conceptual entity, we look first of all for a theory truly describing `the reality of the cognitive realm to which a conceptual entity belongs', and additionally also `how human minds can interact with the latter so that decision-making can occur'.

In this sense, each time we have put forward a model for some specific experimental data, it has always been our preoccupation to also make sure that (i) the model was extracted following the logic that governs our theory of human cognition, and (ii) that whatever other experiments would be performed by a human mind interacting with that same cognitive-conceptual entity under consideration, also the data of these hypothetical additional experiments could have been modeled exactly in the same way. Clearly, this requirement -- that `all possible experiments and data' have to be modeled in an equivalent way -- poses severe constraints to our approach, and it is not a priori evident that this would always be possible. However, we are convinced that the fundamental idea underlying our methodology, namely that of looking upon a decision as an interaction of a human mind with a conceptual entity in a specific state (with such state being independent of the human minds possibly interacting with it), equips the theory of exactly those degrees of freedom that are needed to model `all possible data from all possible experiments'.

As we already explained in the foregoing, in all this we have been guided by how physical theories deal with data coming from the physical domain. They indeed satisfy this criterion and are able to model all data from all possible experiments that can be executed on a given physical entity. What we have called `conceptual entity' is what in physics corresponds to the notion of `physical entity'. Now, in our approach we might be classified as adhering to an idealistic philosophy, i.e. believing that the conceptual entities ``really exist,'' and are not mere creations of our human culture. Our answer to this objection is the following: to profit of the strength of the approach it is not mandatory to take a philosophical stance in the above mentioned way, in the sense that we are not obliged to attribute more existence to what we call a conceptual entity than that attributed, for example, to `human culture' in its entirety. The importance of the approach lies in considering such a conceptual entity as independently existing from any interaction with a human mind, and describe the continuously existing interactions with human minds as processes of the `change of state of the conceptual entity', and whenever applicable also as processes of the `change of context'. And again, let us emphasize that this `hidden-interaction' methodology is inspired by its relevance to physical theories. Our working hypothesis is that in this way it will be possible to advantageously model, and better understand, all of human cognition experimental situations.

Having said this, we observe that the interpretation of the quantum formalism that is commonly used in cognitive domains is a subjectivist one, very similar to that interpretation of quantum theory known as `quantum Bayesianism', or `QBism' \cite{FuchsEtal2014}. In a sense, this interpretation is the polar opposite of our realistic (non-subjectivistic) operational approach. Indeed, QBism originates from a strong critique \cite{fms2014} of the famous Einstein-Podolsky-Rosen reality criterion \cite{epr1935}, whereas at the foundation of the Geneva-Brussels approach there is the idea of taking such criterion not only extremely seriously, but also of using it more thoroughly, as a powerful demarcating tool separating `actually existing properties' from `properties that are only available to be brought into actual existence', and therefore exist in a potential sense \cite{sdb2011}. In other terms,  a quantum state is not considered in QBism as a description of the actual properties of a physical entity, but of the beliefs of the experimenter about it. Similarly, for the majority of authors in quantum cognition, a quantum state is a description of the state of belief of a participant, and not of the actual state of the conceptual entity that interacts with the participants. In ultimate analysis, this difference of perspectives is about taking a clear position regarding the key notion of `certainty': is certainty (probability $1$ assignments) just telling us something about the very firm belief of a subject, or also about some objective properties of the world (be it physical or cultural)? In the same way, are probabilities only shared personal beliefs, based on habit, or also elements of reality (considering that in principle their values can be predicted with certainty)? Although we certainly agree that it is not necessary to take a final stance on these issues to advantageously exploit the quantum mathematics in the modeling of many experimental situations, both in physics and cognition, we also think that the explicative power of a pure subjectivist view rapidly diminishes when we have to address the most remarkable properties of the physical and conceptual entities, like non-locality (non-spatiality) and the non-compositional way with which they can combine. 

It is important to emphasize that the subjectivist view is also a consequence of the absence, in the standard quantum formalism, of a meaningful description of what goes on `behind the scenes' during a measurement. On the other hand, the hidden-measurement paradigm, as implemented in the extended Bloch representation \cite{asdb2014}, or even more generally in the GTR-model \cite{asdb2015b,asdb2015c,asdb2015d}, offers a credible description of the dynamics of a measurement process, in terms of a process of actualization of potential interactions, thus explaining a possible origin of the quantum indeterminism. This certainly allows understanding the so-called `collapse of the state vector' as an objective process, either produced by a macroscopic apparatus in a physics laboratory, or by a mind-brain apparatus in a psychological laboratory. As we tried to motivate in the second part of this article, this completed version of the quantum formalism also allowed us to describe those aspects of a psychological measurements -- the possible different ways participants can choose an outcome -- that would be impossible to model by remaining within the narrow confines not only of the standard formalism, but also of a strict subjectivistic interpretation of it. 

To conclude, a final remark is in order. Quantum cognition is undoubtedly a fascinating field of investigation also for physicists, as it offers the opportunity to take a new look at certain aspects of the quantum formalism and use them to possibly make discoveries also in the physical domain. We already mentioned the example of `entangled measurements', that were necessary to exactly model certain correlations. Entangled (non-separable) measurements are usually not considered in the physics of Bell inequalities, while they are widely explored in quantum cryptography, teleportation and information. However, it is very possible that this stronger form of entanglement will prove to be useful for the interpretation of certain non-locality tests and the explanation of `anomalies' that were identified in EPR-Bell experiments \cite{as2014}. Also, for what concerns the notion of `universal measurement', which is quite natural in psychological measurements, since data are obtained from a collection of different minds, could it be that `universal averages' also happen in the physical domain? In other terms, could it be that a single measurement apparatus is actually more like `a collection of different minds' than `a single Born-like mind'? Considering that the origin of the observed deviations from the Born rule, in situations of sequential measurements, can be understood as the ineffectiveness of the averaging process in producing the Born prescription, is it possible to imagine, in the physics laboratory, similar experimental situations where these deviations would be equally observed, thus confirming that the hypothesis of `hidden measurement-interactions' would be a pertinent one also beyond the psychological domain? Whatever the verdict will be, we certainly live in a very stimulating time for foundational research; a time where the conceptual tools that once helped us building a deeper understanding of the `microscopic layer' of our physical reality are now proving to be instrumental for understanding our human `mental layer'; but also a time where all this is also coming back to physics, not only in the form of possible new experimental findings, but also of possible new and deeper understandings \cite{aerts2009,aerts2010a,aerts2010b,aerts2013,aerts2014}.

\end{document}